\documentclass[10pt,twocolumn,letterpaper]{article}

\usepackage{cvpr}
\usepackage{threeparttable}
\usepackage{bbding}
\usepackage[percent]{overpic}
\usepackage[dvipsnames]{xcolor}
\usepackage{times}
\usepackage{epsfig}
\usepackage{graphicx}
\usepackage{amsmath}
\usepackage{amssymb}
\usepackage{mathtools}
\usepackage{multirow}
\usepackage{enumitem}
\usepackage{caption}
\usepackage{subcaption}
\usepackage{array}
\usepackage{colortbl}
\usepackage{booktabs}
\usepackage{bbm}
\usepackage{multicol}
\usepackage{makecell}
\usepackage{xspace}
\usepackage{float}
\usepackage{color}
\usepackage{lipsum}
\usepackage{pifont}

\usepackage[pagebackref=true,breaklinks=true,letterpaper=true,colorlinks,bookmarks=false]{hyperref}

\cvprfinalcopy 

\def\cvprPaperID{****} 

\newcommand\blfootnote[1]{%
  \begingroup
  \renewcommand\thefootnote{}\footnote{#1}%
  \addtocounter{footnote}{-1}%
  \endgroup
}
\begin{document}

\title{YOLOX: Exceeding YOLO Series in 2021}

\author{Zheng Ge$^*$~~~ Songtao Liu$^{*\dag}$~~~ Feng Wang~~~ Zeming Li~~~ Jian Sun\\
	Megvii Technology\\
	\tt\small \{gezheng, liusongtao, wangfeng02, lizeming, sunjian\}@megvii.com}

\twocolumn[{
\renewcommand\twocolumn[1][]{#1}
\maketitle
\vspace{-11mm}
\begin{figure}[H]
\hsize=\textwidth
\centering
\begin{subfigure}{0.45\textwidth}
    \centering
    \includegraphics[width=1\textwidth]{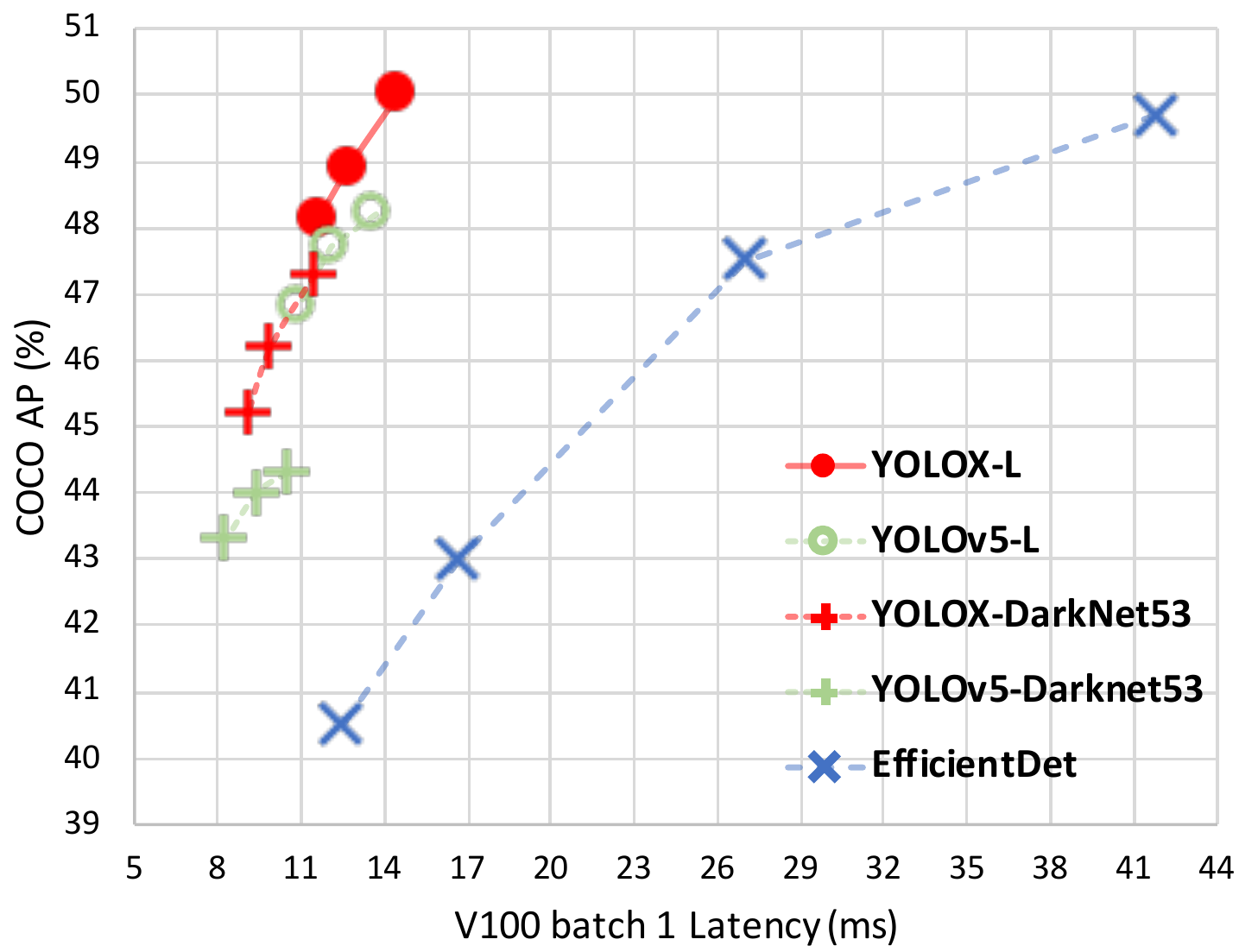}
    \label{fig:1a}	
\end{subfigure}    
\hspace{0.2in}
\begin{subfigure}{0.48\textwidth}
     \centering
     \includegraphics[width=1\textwidth]{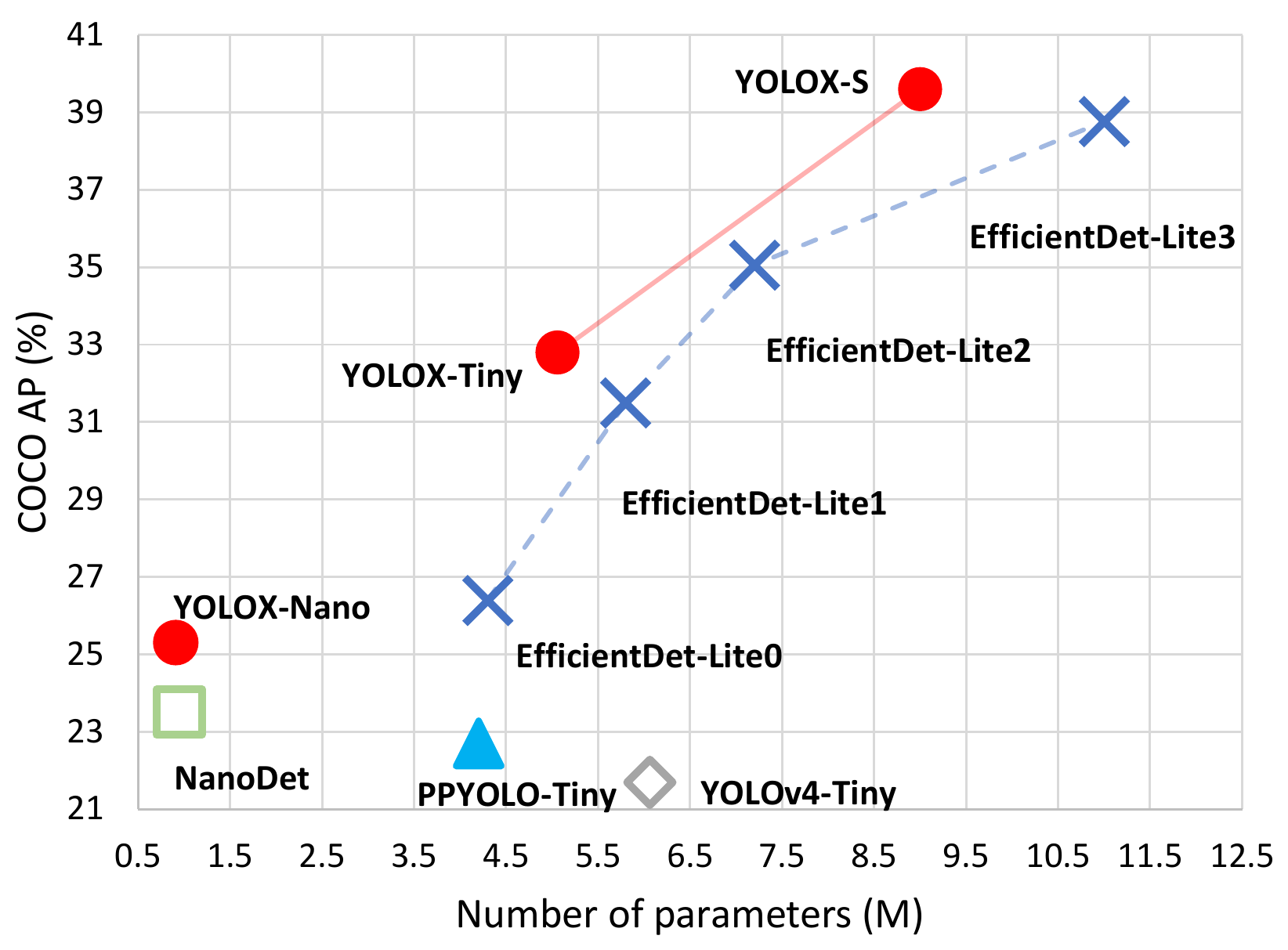}
     \label{fig:1b}
\end{subfigure}
\hspace{0.in}
\vspace{-6mm}
\caption{Speed-accuracy trade-off of accurate models (top) and Size-accuracy curve of lite models on mobile devices (bottom) for YOLOX and other state-of-the-art object detectors.}
\label{fig:speed}
\end{figure}
\vspace{2mm}
}]

\blfootnote{* Equal contribution.}
\blfootnote{\dag~Corresponding author.}

\begin{abstract}
	In this report, we present some experienced improvements to YOLO series, forming a new high-performance detector\,---\,YOLOX. We switch the YOLO detector to an anchor-free manner and conduct other advanced detection techniques, \emph{i.e.}, a decoupled head and the leading label assignment strategy SimOTA to achieve state-of-the-art results across a large scale range of models: For YOLO-Nano with only 0.91M parameters and 1.08G FLOPs, we get 25.3\% AP on COCO, surpassing NanoDet by 1.8\% AP; for YOLOv3, one of the most widely used detectors in industry, we boost it to 47.3\% AP on COCO, outperforming the current best practice by 3.0\% AP; for YOLOX-L with roughly the same amount of parameters as YOLOv4-CSP, YOLOv5-L, we achieve 50.0\% AP on COCO at a speed of 68.9 FPS on Tesla V100, exceeding YOLOv5-L by 1.8\% AP. Further, we won the 1st Place on Streaming Perception Challenge (Workshop on Autonomous Driving at CVPR 2021) using a single YOLOX-L model. We hope this report can provide useful experience for developers and researchers in practical scenes, and we also provide deploy versions with ONNX, TensorRT, NCNN, and Openvino supported. Source code is at \url{https://github.com/Megvii-BaseDetection/YOLOX}.      
\end{abstract}


\section{Introduction}

With the development of object detection, YOLO series~\cite{yolo,yolo9000,yolov3,yolov4,yolov5} always pursuit the optimal speed and accuracy trade-off for real-time applications. They extract the most advanced detection technologies available at the time (\emph{e.g.}, anchors~\cite{FasterCNN} for YOLOv2~\cite{yolo9000}, Residual Net~\cite{ResNet} for YOLOv3~\cite{yolov3}) and optimize the implementation for best practice. Currently, YOLOv5~\cite{yolov5} holds the best trade-off performance with 48.2\% AP on COCO at 13.7 ms.\footnote{we choose the YOLOv5-L model at $640\times640$ resolution and test the model with FP16-precision and batch=1 on a V100 to align the settings of YOLOv4~\cite{yolov4} and YOLOv4-CSP~\cite{scaleyolo} for a fair comparison}   

Nevertheless, over the past two years, the major advances in object detection academia have focused on anchor-free detectors~\cite{fcos,centernet,cornernet}, advanced label assignment strategies~\cite{freeanchor,atss,paa,autoassign,iqdet,ota}, and end-to-end (NMS-free) detectors~\cite{detr,end2end,end2end2}. These have not been integrated into YOLO families yet, as YOLOv4 and YOLOv5 are still anchor-based detectors with hand-crafted assigning rules for training.

That's what brings us here, delivering those recent advancements to YOLO series with experienced optimization. Considering YOLOv4 and YOLOv5 may be a little over-optimized for the anchor-based pipeline, we choose YOLOv3~\cite{yolov3} as our start point (we set YOLOv3-SPP as the default YOLOv3). Indeed, YOLOv3 is still one of the most widely used detectors in the industry due to the limited computation resources and the insufficient software support in various practical applications.  

\begin{figure*}[thbp]
	\begin{center}
		\includegraphics[width=0.77\linewidth]{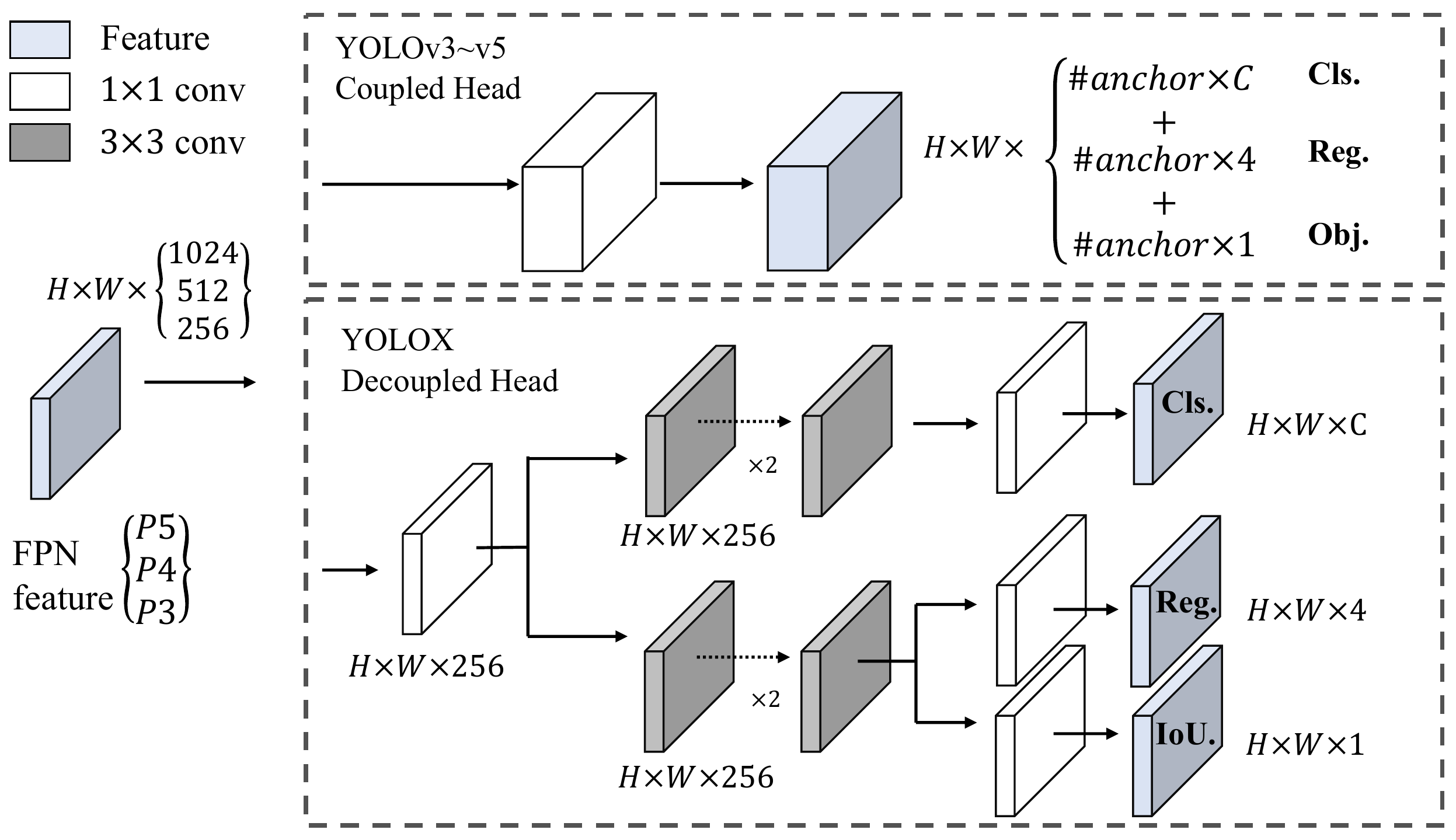}
	\end{center}
	\caption{Illustration of the difference between YOLOv3 head and the proposed decoupled head. For each level of FPN feature, we first adopt a $1\times1$ conv layer to reduce the feature channel to 256 and then add two parallel branches with two $3\times3$ conv layers each for classification and regression tasks respectively. IoU branch is added on the regression branch. }
	\label{fig:head1}
\end{figure*}

As shown in Fig.~\ref{fig:speed}, with the experienced updates of the above techniques, we boost the YOLOv3 to 47.3\% AP (YOLOX-DarkNet53) on COCO with $640\times640$ resolution, surpassing the current best practice of YOLOv3 (44.3\% AP, ultralytics version\footnote{\url{https://github.com/ultralytics/yolov3}\label{yolov3-ultra}}) by a large margin. Moreover, when switching to the advanced YOLOv5 architecture that adopts an advanced CSPNet~\cite{cspnet} backbone and an additional PAN~\cite{pan} head, YOLOX-L achieves 50.0\% AP on COCO with $640\times640$ resolution, outperforming the counterpart YOLOv5-L by 1.8\% AP. We also test our design strategies on models of small size. YOLOX-Tiny and YOLOX-Nano (only 0.91M Parameters and 1.08G FLOPs) outperform the corresponding counterparts YOLOv4-Tiny and NanoDet\footnote{\url{https://github.com/RangiLyu/nanodet}\label{nano}} by 10\% AP and 1.8\% AP, respectively.

We have released our code at \url{https://github.com/Megvii-BaseDetection/YOLOX}, with ONNX, TensorRT, NCNN and Openvino supported. One more thing worth mentioning, we won the 1st Place on Streaming Perception Challenge (Workshop on Autonomous Driving at CVPR 2021) using a single YOLOX-L model.

\section{YOLOX}
\subsection{YOLOX-DarkNet53}
We choose YOLOv3~\cite{yolov3} with Darknet53 as our baseline. In the following part, we will walk through the whole system designs in YOLOX step by step.

\paragraph{Implementation details}\label{settings} Our training settings are mostly consistent from the baseline to our final model. We train the models for a total of 300 epochs with 5 epochs warm-up on COCO \textit{train2017}~\cite{MSCOCO}. We use stochastic gradient descent (SGD) for training. We use a learning rate of $lr\times$BatchSize/64 (linear scaling~\cite{linear}), with a initial $lr =$ 0.01 and the cosine lr schedule. The weight decay is 0.0005 and the SGD momentum is 0.9. The batch size is 128 by default to typical 8-GPU devices. Other batch sizes include single GPU training also work well. The input size is evenly drawn from 448 to 832 with 32 strides. FPS and latency in this report are all measured with FP16-precision and batch=1 on a single Tesla V100.

\paragraph{YOLOv3 baseline} Our baseline adopts the architecture of DarkNet53 backbone and an SPP layer, referred to YOLOv3-SPP in some papers~\cite{yolov4,yolov5}. We slightly change some training strategies compared to the original implementation~\cite{yolov3}, adding EMA weights updating, cosine lr schedule, IoU loss and IoU-aware branch. We use BCE Loss for training \emph{cls} and \emph{obj} branch, and IoU Loss for training \emph{reg} branch. These general training tricks are orthogonal to the key improvement of YOLOX, we thus put them on the baseline. Moreover, we only conduct \texttt{RandomHorizontalFlip}, \texttt{ColorJitter} and multi-scale for data augmentation and discard the \texttt{RandomResizedCrop} strategy, because we found the \texttt{RandomResizedCrop} is kind of overlapped with the planned mosaic augmentation. With those enhancements, our baseline achieves 38.5\% AP on COCO \textit{val}, as shown in Tab.~\ref{table:yolov3}.

\paragraph{Decoupled head} In object detection, the conflict between classification and regression tasks is a well-known problem~\cite{TSD,doubleheadrcnn}. Thus the decoupled head for  classification and localization is widely used in the most of one-stage and two-stage detectors~\cite{focal-loss,fcos,decouplehead,doubleheadrcnn}. However, as YOLO series' backbones and feature pyramids (~\emph{e.g.}, FPN~\cite{pfp}, PAN~\cite{panet}.) continuously evolving, their detection heads remain coupled as shown in Fig.~\ref{fig:head1}. 

\begin{figure}[t]
	\centering
	\includegraphics[width=0.99\linewidth]{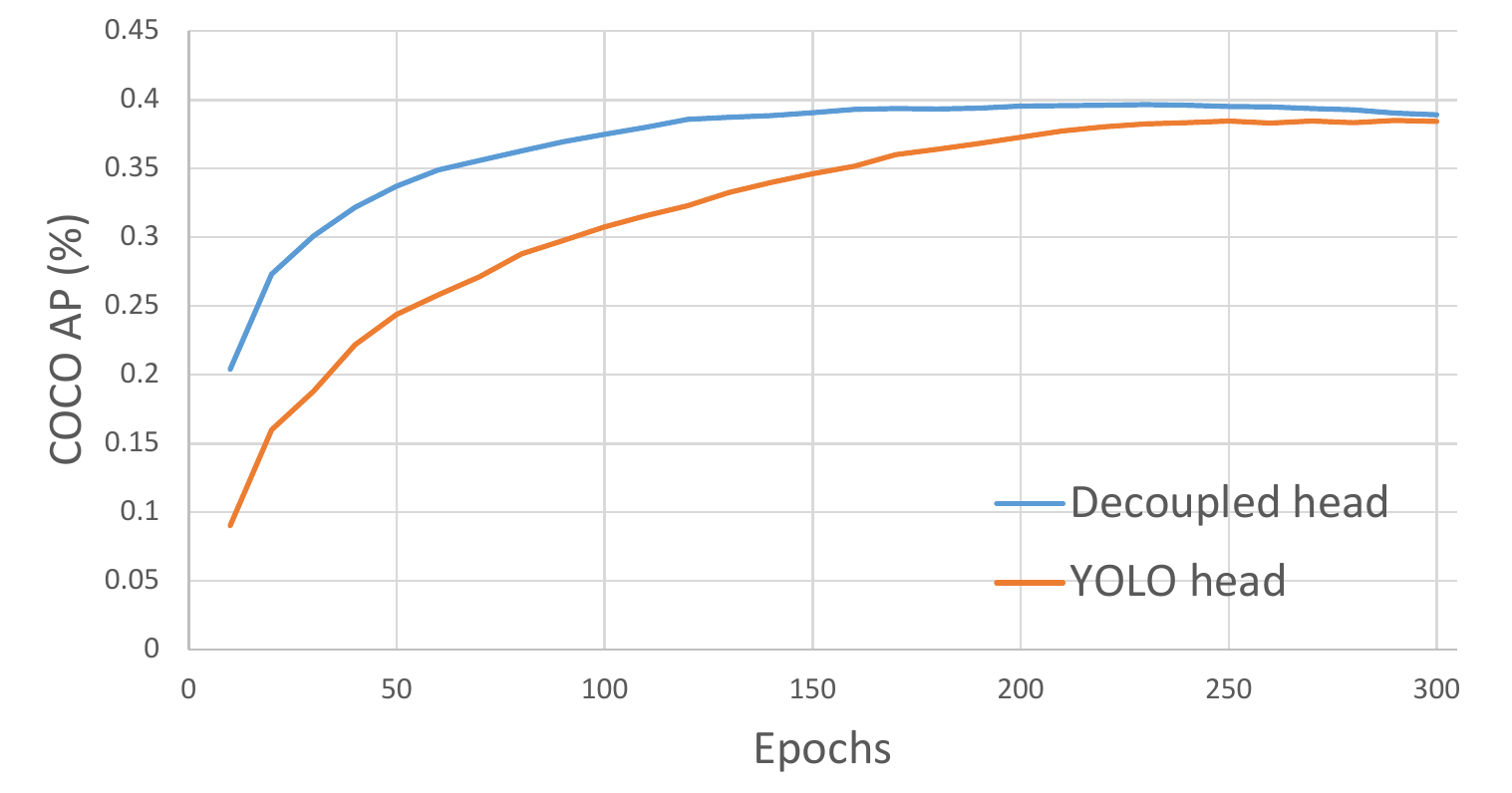}
	\caption{Training curves for detectors with YOLOv3 head or decoupled head. We evaluate the AP on COCO \emph{val} every 10 epochs. It is obvious that the decoupled head converges much faster than the YOLOv3 head and achieves better result finally.}
	\label{fig:head2}
\end{figure}

Our two analytical experiments indicate that the coupled detection head may harm the performance. 1). Replacing YOLO's head with a decoupled one greatly improves the converging speed as shown in Fig.~\ref{fig:head2}. 2). The decoupled head is essential to the end-to-end version of YOLO (will be described next). One can tell from Tab.~\ref{e2edec}, the end-to-end property decreases by 4.2\% AP with the coupled head, while the decreasing reduces to 0.8\% AP for a decoupled head. We thus replace the YOLO detect head with a lite decoupled head as in Fig.~\ref{fig:head1}. Concretely, it contains a $1\times1$ conv layer to reduce the channel dimension, followed by two parallel branches with two $3\times3$ conv layers respectively. We report the inference time with batch=1 on V100 in Tab.~\ref{table:yolov3} and the lite decoupled head brings additional 1.1 ms (11.6 ms ~\emph{v.s.} 10.5 ms). 

\begin{table}[!t]
\centering
\begin{threeparttable}
\begin{tabular}{l|c|c}
\toprule
Models       & Coupled Head & Decoupled Head \\ \midrule
Vanilla YOLO & 38.5         & 39.6           \\
End-to-end YOLO & 34.3 \color{RubineRed}{\small \textbf{(-4.2)}} & 38.8 \color{RubineRed}{\small \textbf{(-0.8)}}\\ \bottomrule
\end{tabular}
\end{threeparttable}
	\caption{The effect of decoupled head for end-to-end YOLO in terms of AP (\%) on COCO.}\label{e2edec}
\end{table}

\paragraph{Strong data augmentation} We add Mosaic and MixUp into our augmentation strategies to boost YOLOX's performance. Mosaic is an efficient augmentation strategy proposed by ultralytics-YOLOv3\textsuperscript{\ref{yolov3-ultra}}. It is then widely used in YOLOv4~\cite{yolov4}, YOLOv5~\cite{yolov5} and other detectors~\cite{yolof}. MixUp~\cite{mixup} is originally designed for image classification task but then modified in BoF~\cite{bag} for object detection training. We adopt the MixUp and Mosaic implementation in our model and close it for the last 15 epochs, achieving 42.0\% AP in Tab.~\ref{table:yolov3}.  After using strong data augmentation, we found ImageNet pre-training is no more beneficial, \textbf{we thus train all the following models from scratch.}

\begin{table*}[!thbp]
	\begin{center}
		\scalebox{0.7}{
			\resizebox{\textwidth}{!}{
				\begin{threeparttable}
					\begin{tabular}{l|l|l|l|l|l}
						\toprule
						Methods              & AP (\%) & Parameters & GFLOPs & Latency & FPS  \\ \midrule
						YOLOv3-ultralytics\textsuperscript{\ref{yolov3-ultra}}   & 44.3   & 63.00 M    & 157.3  & 10.5 ms    & 95.2 \\ \midrule
						YOLOv3 baseline      & 38.5   & 63.00 M    & 157.3  & 10.5 ms    & 95.2 \\ 
						\ \ \ \ \small{+decoupled head}    & 39.6 \color{ForestGreen}\small \textbf{(+1.1)}   & 63.86 M    & 186.0  & 11.6 ms    & 86.2 \\
						\ \ \ \ \small{+strong augmentation} & 42.0 \color{ForestGreen}\small \textbf{(+2.4)}  & 63.86 M    & 186.0  & 11.6 ms    & 86.2 \\ 
						\ \ \ \ \small{+anchor-free}       & 42.9 \color{ForestGreen}\small \textbf{(+0.9)}   & 63.72 M    & 185.3  & 11.1 ms    & 90.1 \\
						\ \ \ \ \small{+multi positives}     & 45.0 \color{ForestGreen}\small \textbf{(+2.1)}  & 63.72 M    & 185.3  & 11.1 ms    & 90.1 \\
						\ \ \ \ \small{+SimOTA}              & \textbf{47.3} \color{ForestGreen}\small \textbf{(+2.3)}  & 63.72 M    & 185.3  & 11.1 ms    & 90.1 \\ 
						\textcolor[RGB]{128,138,135}{\ \ \ \ \small{+NMS free (optional)}}            & \textcolor[RGB]{128,138,135}{46.5}  \small \textcolor[RGB]{128,138,135}{(-0.8)}  & \textcolor[RGB]{128,138,135}{67.27 M}    & \textcolor[RGB]{128,138,135}{205.1}  & \textcolor[RGB]{128,138,135}{13.5 ms}    & \textcolor[RGB]{128,138,135}{74.1} \\ \bottomrule
						\end{tabular}
		\end{threeparttable}}}
	\end{center}
	\caption{Roadmap of YOLOX-Darknet53 in terms of AP (\%) on COCO \emph{val}. All the models are tested at $640\times640$ resolution, with FP16-precision and batch=1 on a Tesla V100. The latency and FPS in this table are measured without post-processing.}
	\label{table:yolov3}
\end{table*}

\paragraph{Anchor-free} Both YOLOv4~\cite{yolov4} and YOLOv5~\cite{yolov5} follow the original anchor-based pipeline of YOLOv3~\cite{yolov3}. However, the anchor mechanism has many known problems. First, to achieve optimal detection performance, one needs to conduct clustering analysis to determine a set of optimal anchors before training. Those clustered anchors are domain-specific and less generalized. Second, anchor mechanism increases the complexity of detection heads, as well as the number of predictions for each image. On some edge AI systems, moving such large amount of predictions between devices (\emph{e.g.}, from NPU to CPU) may become a potential bottleneck in terms of the overall latency.

Anchor-free detectors~\cite{fcos,centernet,cornernet} have developed rapidly in the past two year. These works have shown that the performance of anchor-free detectors can be on par with anchor-based detectors. Anchor-free mechanism significantly reduces the number of design parameters which need heuristic tuning and many tricks involved (\emph{e.g.}, Anchor Clustering~\cite{yolo9000}, Grid Sensitive~\cite{ppyolo}.) for good performance, making the detector, especially its training and decoding phase, \textit{considerably} simpler~\cite{fcos}. 

Switching YOLO to an anchor-free manner is quite simple. We reduce the predictions for each location from 3 to 1 and make them directly predict four values, \emph{i.e.}, two offsets in terms of the left-top corner of the grid, and the height and width of the predicted box. We assign the center location of each object as the positive sample and pre-define a scale range, as done in~\cite{fcos}, to designate the FPN level for each object. Such modification reduces the parameters and GFLOPs of the detector and makes it faster, but obtains better performance -- 42.9\% AP as shown in Tab.~\ref{table:yolov3}. 

\paragraph{Multi positives} To be consistent with the assigning rule of YOLOv3, the above anchor-free version selects only ONE positive sample (the center location) for each object meanwhile ignores other high quality predictions. However, optimizing those high quality predictions may also bring beneficial gradients, which may alleviates the extreme imbalance of positive/negative sampling during training. We simply assigns the center $3\times3$ area as positives, also named ``center sampling'' in FCOS~\cite{fcos}. The performance of the detector improves to 45.0\% AP as in Tab.~\ref{table:yolov3}, already surpassing the current best practice of ultralytics-YOLOv3 (44.3\% AP\textsuperscript{\ref{yolov3-ultra}}).     

\paragraph{SimOTA} Advanced label assignment is another important progress of object detection in recent years. Based on our own study OTA~\cite{ota}, we conclude four key insights for an advanced label assignment: 1). loss/quality aware, 2). center prior, 3). dynamic number of positive anchors\footnote{The term ``anchor'' refers to ``anchor point'' in the context of anchor-free detectors and ``grid'' in the context of YOLO.} for each ground-truth (abbreviated as dynamic top-k), 4). global view. OTA meets all four rules above, hence we choose it as a candidate label assigning strategy.

Specifically, OTA~\cite{ota} analyzes the label assignment from a global perspective and formulate the assigning procedure as an Optimal Transport (OT) problem, producing the SOTA performance among the current assigning strategies~\cite{paa,autoassign,atss,iqdet,freeanchor}. However, in practice we found solving OT problem via Sinkhorn-Knopp algorithm brings 25\% extra training time, which is quite expensive for training 300 epochs. We thus simplify it to dynamic top-k strategy, named SimOTA, to get an approximate solution. 

We briefly introduce SimOTA here. SimOTA first calculates pair-wise matching degree, represented by cost~\cite{ota, lla, paa, detr} or quality~\cite{defcn} for each prediction-\emph{gt} pair. For example, in SimOTA, the cost between gt $g_i$ and prediction $p_j$ is calculated as:
\begin{equation}
\begin{split}
c_{ij} = & L_{ij}^{cls} + \lambda L_{ij}^{reg},
\end{split} \label{cost}
\end{equation}
where $\lambda$ is a balancing coefficient. $L_{ij}^{cls}$ and $L_{ij}^{reg}$ are classficiation loss and regression loss between gt $g_i$ and prediction $p_j$. Then, for gt $g_i$, we select the top $k$ predictions with the least cost within a fixed center region as its positive samples. Finally, the corresponding grids of those positive predictions are assigned as positives, while the rest grids are negatives. 
Noted that the value $k$ varies for different ground-truth. Please refer to Dynamic $k$ Estimation strategy in OTA~\cite{ota} for more details.

SimOTA not only reduces the training time but also avoids additional solver hyperparameters in Sinkhorn-Knopp algorithm. As shown in Tab.~\ref{table:yolov3}, SimOTA raises the detector from 45.0\% AP to 47.3\% AP,  higher than the SOTA ultralytics-YOLOv3 by 3.0\% AP, showing the power of the advanced assigning strategy.

\paragraph{End-to-end YOLO} We follow \cite{end2end2} to add two additional conv layers, one-to-one label assignment, and stop gradient. These enable the detector to perform an end-to-end manner, but slightly decreasing the performance and the inference speed, as listed in Tab.~\ref{table:yolov3}. We thus leave it as an optional module which is not involved in our final models.      

\begin{table}[!thbp]
	\begin{center}
		\scalebox{0.47}{
			\resizebox{\textwidth}{!}{
				\begin{threeparttable}
					\begin{tabular}{l|l|l|l|l}
						\toprule
						Models   & AP (\%) & Parameters & GFLOPs & Latency \\ \midrule
						YOLOv5-S & 36.7   & 7.3 M      & 17.1   & 8.7 ms  \\
						YOLOX-S  & \textbf{39.6} \color{ForestGreen}\small \textbf{(+2.9)}   & 9.0 M      & 26.8   & 9.8 ms  \\ \midrule
						YOLOv5-M & 44.5   & 21.4 M     & 51.4   & 11.1 ms \\
						YOLOX-M  & \textbf{46.4} \color{ForestGreen}\small \textbf{(+1.9)}   & 25.3 M     & 73.8   & 12.3 ms \\ \midrule
						YOLOv5-L & 48.2   & 47.1 M     & 115.6  & 13.7 ms \\
						YOLOX-L  & \textbf{50.0} \color{ForestGreen}\small \textbf{(+1.8)}   & 54.2 M     & 155.6  & 14.5 ms \\ \midrule
						YOLOv5-X & 50.4   & 87.8 M     & 219.0  & 16.0 ms \\
						YOLOX-X  & \textbf{51.2} \color{ForestGreen}\small \textbf{(+0.8)}   & 99.1 M     & 281.9  & 17.3 ms \\ \bottomrule
					\end{tabular}
		\end{threeparttable}}}
	\end{center}
	\caption{Comparison of YOLOX and YOLOv5 in terms of AP (\%) on COCO. All the models are tested at $640\times640$ resolution, with FP16-precision and batch=1 on a Tesla V100.}
	\label{table:yolov5}
\end{table}

\begin{table}[!thbp]
	\begin{center}
		\scalebox{0.42}{
			\resizebox{\textwidth}{!}{
				\begin{threeparttable}
					\begin{tabular}{l|l|c|c}
						\toprule
						Models      & AP (\%) & Parameters & GFLOPs \\ \midrule
						YOLOv4-Tiny~\cite{scaleyolo} & 21.7   & 6.06 M      & 6.96   \\
						PPYOLO-Tiny  & 22.7    & 4.20 M      & -   \\ 
						YOLOX-Tiny  & \textbf{32.8} \color{ForestGreen}\small \textbf{(+10.1)}   & 5.06 M      & 6.45    \\ \midrule
						NanoDet\textsuperscript{\ref{nano}}     & 23.5   & 0.95 M      & 1.20    \\
						YOLOX-Nano  & \textbf{25.3} \color{ForestGreen}\small \textbf{(+1.8)}   & 0.91 M      & 1.08    \\ \bottomrule
					\end{tabular}
		\end{threeparttable}}}
	\end{center}
	\caption{Comparison of YOLOX-Tiny and YOLOX-Nano and the counterparts in terms of AP (\%) on COCO \textit{val}. All the models are tested at $416\times416$ resolution.}
	\label{table:yolotiny}
\end{table}

\subsection{Other Backbones}
Besides DarkNet53, we also test YOLOX on other backbones with different sizes, where YOLOX achieves consistent improvements against all the corresponding counterparts.   

\paragraph{Modified CSPNet in YOLOv5} To give a fair comparison, we adopt the exact YOLOv5's backbone including modified CSPNet~\cite{cspnet}, SiLU activation, and the PAN~\cite{pan} head. We also follow its scaling rule to product YOLOX-S, YOLOX-M, YOLOX-L, and YOLOX-X models. Compared to YOLOv5 in Tab.~\ref{table:yolov5}, our models get consistent improvement by $\sim$3.0\% to $\sim$1.0\% AP, with only marginal time increasing (comes from the decoupled head). 
\paragraph{Tiny and Nano detectors} We further shrink our model as YOLOX-Tiny to compare with YOLOv4-Tiny~\cite{scaleyolo}. For mobile devices, we adopt depth wise convolution to construct a YOLOX-Nano model, which has only 0.91M parameters and 1.08G FLOPs. As shown in Tab.~\ref{table:yolotiny}, YOLOX performs well with even smaller model size than the counterparts.  

\paragraph{Model size and data augmentation} In our experiments, all the models keep almost the same learning schedule and optimizing parameters as depicted in~\ref{settings}. However, we found that the suitable augmentation strategy varies across different size of models. As Tab.~\ref{aug} shows, while applying MixUp for YOLOX-L can improve AP by 0.9\%, it is better to weaken the augmentation for small models like YOLOX-Nano. Specifically, we remove the mix up augmentation and weaken the mosaic (reduce the scale range from [0.1, 2.0] to [0.5, 1.5]) when training small models, \emph{i.e.}, YOLOX-S, YOLOX-Tiny, and YOLOX-Nano. Such a modification improves YOLOX-Nano's AP from 24.0\% to 25.3\%. 

For large models, we also found that stronger augmentation is more helpful. Indeed, our MixUp implementation is part of heavier than the original version in ~\cite{bag}. Inspired by Copypaste~\cite{copypaste}, we jittered both images by a random sampled scale factor before mixing up them. To understand the power of Mixup with scale jittering, we compare it with Copypaste on YOLOX-L. Noted that Copypaste requires extra instance mask annotations while MixUp does not. But as shown in Tab.~\ref{aug}, these two methods achieve competitive performance, indicating that MixUp with scale jittering is a qualified replacement for Copypaste when no instance mask annotation is available.



\begin{table}[!thbp]
\centering
\scalebox{0.42}{
\resizebox{\textwidth}{!}{
\begin{threeparttable}
\begin{tabular}{l|c|c|l}
\toprule
Models                      & Scale Jit. & Extra Aug.      & AP (\%) \\ \midrule
\multirow{2}{*}{YOLOX-Nano} & [0.5, 1.5]             &   -       &   25.3    \\
                            & [0.1, 2.0]             & MixUp     &   24.0 \color{RubineRed}{\small \textbf{(-1.3)}}    \\ \midrule
\multirow{3}{*}{YOLOX-L}    & [0.1, 2.0]             & -         &   48.6    \\
                            & [0.1, 2.0]             & MixUp     &   49.5 \color{ForestGreen}\small \textbf{(+0.9)}    \\ \cmidrule{2-4}
                            & [0.1, 2.0]             & Copypaste~\cite{copypaste} &   49.4 \\ \bottomrule
\end{tabular}
\end{threeparttable}}}
\caption{Effects of data augmentation under different model sizes. ``Scale Jit.'' stands for the range of scale jittering for mosaic image. Instance mask annotations from COCO $trainval$ are used when adopting Copypaste.}\label{aug}
\end{table}

\section{Comparison with the SOTA}
There is a tradition to show the SOTA comparing table as in Tab.~\ref{table:sota}. However, keep in mind that the inference speed of the models in this table is often uncontrolled, as speed varies with software and hardware. We thus use the same hardware and code base for all the YOLO series in Fig.~\ref{fig:speed}, plotting the somewhat controlled speed/accuracy curve.

We notice that there are some high performance YOLO series with larger model sizes like Scale-YOLOv4~\cite{scaleyolo} and YOLOv5-P6~\cite{yolov5}. And the current Transformer based detectors~\cite{swin} push the accuracy-SOTA to $\sim$60 AP. Due to the time and resource limitation, we did not explore those important features in this report. However, they are already in our scope.

\begin{table*}[h]
	\centering
	\resizebox{0.9\textwidth}{!}{
		\begin{tabular}{l|l|c|c|c|cccccc}
			\toprule
			\multirow{2}{*}{\textbf{Method}} & \multirow{2}{*}{\textbf{Backbone}} & \multirow{2}{*}{\textbf{Size}} &\multirow{2}{*}{\textbf{FPS}} &
			\multirow{2}{*}{\textbf{AP (\%)}} & \multirow{2}{*}{\textbf{AP$_{50}$}} & \multirow{2}{*}{\textbf{AP$_{75}$}} & \multirow{2}{*}{\textbf{AP$_S$}} & \multirow{2}{*}{\textbf{AP$_M$}} & \multirow{2}{*}{\textbf{AP$_L$}}\\
			& & & \tiny{(V100)} & & & & & & &	 \\			
			\hline
			\midrule
			YOLOv3 + ASFF* \cite{asff}& Darknet-53 & 608 &  45.5    &42.4  & 63.0  & 47.4  & 25.5  & 45.7  & 52.3  \\
			YOLOv3 + ASFF* \cite{asff} & Darknet-53 & 800 & 29.4    & 43.9  & 64.1  & 49.2  & 27.0  & 46.6  & 53.4  \\
			\midrule
			EfficientDet-D0~\cite{efficientdet} & Efficient-B0  & 512 &  98.0  & 33.8  & 52.2  & 35.8  & 12.0  & 38.3  & 51.2  \\
			EfficientDet-D1~\cite{efficientdet} & Efficient-B1 & 640 &  74.1  & 39.6  & 58.6  & 42.3  & 17.9  & 44.3  & 56.0  \\
			EfficientDet-D2~\cite{efficientdet} & Efficient-B2 & 768 & 56.5  & 43.0  & 62.3  & 46.2  & 22.5  & 47.0  & 58.4  \\
			EfficientDet-D3~\cite{efficientdet} & Efficient-B3 & 896 & 34.5   & 45.8  & 65.0  & 49.3  & 26.6  & 49.4  & 59.8  \\
			\midrule
			PP-YOLOv2~\cite{ppyolo} & ResNet50-vd-dcn & 640 & 68.9 &  49.5  & 68.2  & 54.4  & 30.7  & 52.9  & 61.2  \\
			PP-YOLOv2~\cite{ppyolo}& ResNet101-vd-dcn & 640 & 50.3 &   50.3  & 69.0  & 55.3  & 31.6  & 53.9  & 62.4  \\
			\midrule
			YOLOv4~\cite{yolov4} & CSPDarknet-53 & 608  & 62.0    & 43.5  & 65.7  & 47.3  & 26.7  & 46.7  & 53.3  \\
			YOLOv4-CSP~\cite{scaleyolo} & Modified CSP & 640 & 73.0   & 47.5  & 66.2  & 51.7  & 28.2  & 51.2  & 59.8  \\
			\midrule
			YOLOv3-ultralytics\textsuperscript{\ref{yolov3-ultra}} & Darknet-53 & 640 & 95.2 & 44.3   & 64.6  & - & - & - & - \\
			YOLOv5-M~\cite{yolov5} & Modified CSP v5 & 640 & 90.1 & 44.5   & 63.1  & - & - & - & - \\
			YOLOv5-L~\cite{yolov5} & Modified CSP v5 & 640 & 73.0 & 48.2   & 66.9  & - & - & - & - \\
			YOLOv5-X~\cite{yolov5} & Modified CSP v5 & 640 & 62.5  & 50.4  & 68.8  & - & - & - & - \\
			\midrule
			YOLOX-DarkNet53       & Darknet-53 & 640 & 90.1 & 47.4  & 67.3  & 52.1 & 27.5 & 51.5 & 60.9 \\
			YOLOX-M        		  & Modified CSP v5 & 640 & 81.3 & 46.4  & 65.4  & 50.6 & 26.3 & 51.0 & 59.9 \\
			YOLOX-L               & Modified CSP v5 & 640 & 69.0 & 50.0   & 68.5  & 54.5 & 29.8 & 54.5 & 64.4 \\
			YOLOX-X               & Modified CSP v5 & 640 & 57.8 & \textbf{51.2}  & 69.6  & 55.7 & 31.2 & 56.1 & 66.1 \\
			\bottomrule
		\end{tabular}
	}
	\caption{Comparison of the speed and accuracy of different object detectors on COCO 2017 \emph{test-dev}. We select all the models trained on 300 epochs for fair comparison.
	}
	\label{table:sota}
\end{table*}

\section{1st Place on Streaming Perception Challenge (WAD at CVPR 2021)}
Streaming Perception Challenge on WAD 2021 is a joint evaluation of accuracy and latency through a recently proposed metric: streaming accuracy~\cite{stream}. The key insight behind this metric is to jointly evaluate the output of the entire perception stack at every time instant, forcing the stack to consider the amount of streaming data that should be ignored while computation is occurring~\cite{stream}. We found that the best trade-off point for the metric on 30 FPS data stream is a powerful model with the inference time $\leq$ 33ms. So we adopt a YOLOX-L model with TensorRT to product our final model for the challenge to win the 1st place. Please refer to the challenge website\footnote{\url{https://eval.ai/web/challenges/challenge-page/800/overview}} for more details.    

\section{Conclusion}
In this report, we present some experienced updates to YOLO series, which forms a high-performance anchor-free detector called YOLOX. Equipped with some recent advanced detection techniques, \emph{i.e.}, decoupled head, anchor-free, and advanced label assigning strategy, YOLOX achieves a better trade-off between speed and accuracy than other counterparts across all model sizes. It is remarkable that we boost the architecture of YOLOv3, which is still one of the most widely used detectors in industry due to its broad compatibility, to 47.3\% AP on COCO, surpassing the current best practice by 3.0\% AP. We hope this report can help developers and researchers get better experience in practical scenes.    

\section*{Acknowledge}  
This research was supported by National Key R\&D Program of China (No. 2017YFA0700800). It was also funded by China
Postdoctoral Science Foundation (2021M690375) and Beijing Postdoctoral Research Foundation          

{\small
\bibliographystyle{ieee_fullname}
\bibliography{egbib}
}

\end{document}